\documentclass[11pt]{article}
\usepackage{graphicx}
\usepackage{amssymb}
\usepackage{amscd}
\usepackage{mathrsfs}
\usepackage{longtable,lscape}
\usepackage{amsthm}
\usepackage{amsfonts}
\usepackage{amsmath}
\usepackage{bbm}
\usepackage{float}
\usepackage{url}

\newcommand{\captionfonts}{\footnotesize}
\makeatletter  
\long\def\@makecaption#1#2{%
  \vskip\abovecaptionskip
  \sbox\@tempboxa{{\captionfonts #1: #2}}%
  \ifdim \wd\@tempboxa >\hsize
    {\captionfonts #1: #2\par}
  \else
    \hbox to\hsize{\hfil\box\@tempboxa\hfil}%
  \fi
  \vskip\belowcaptionskip}
\makeatother 
\begin{document}
\title{Quantum Cognition Beyond Hilbert Space I: Fundamentals}
\author{Diederik Aerts$^1$, Lyneth Beltran$^1$, Massimiliano Sassoli de Bianchi$^{2}$, Sandro Sozzo$^{3}$ and Tomas Veloz$^{1}$ \vspace{0.5 cm} \\ 
        \normalsize\itshape
        $^1$ Center Leo Apostel for Interdisciplinary Studies, 
         Brussels Free University \\ 
        \normalsize\itshape
         Krijgskundestraat 33, 1160 Brussels, Belgium \\
        \normalsize
        E-Mails: \url{diraerts@vub.ac.be,lyneth.benedictinelawcenter@gmail.com}
          \vspace{0.5 cm} \\ 
        \normalsize\itshape
        $^2$ Laboratorio di Autoricerca di Base \\
        \normalsize\itshape
        6914 Lugano, Switzerland
        \\
        \normalsize
        E-Mail: \url{autoricerca@gmail.com}
          \vspace{0.5 cm} \\ 
        \normalsize\itshape
        $^3$ School of Management and IQSCS, University of Leicester \\ 
        \normalsize\itshape
         University Road, LE1 7RH Leicester, United Kingdom \\
        \normalsize
        E-Mail: \url{ss831@le.ac.uk} 
       	\\
              }
\date{}
\maketitle
\begin{abstract}
\noindent
The formalism of quantum theory in Hilbert space has been applied with success to the modeling and explanation of several cognitive phenomena, whereas traditional cognitive approaches were problematical. However, this `quantum cognition paradigm' was recently challenged by its proven impossibility to simultaneously model `question order effects' and `response replicability'. In Part I of this paper we describe sequential dichotomic measurements within 
an operational and realistic framework for human cognition elaborated by ourselves, and represent them in a quantum-like `extended Bloch representation' where the Born rule of quantum probability does not necessarily
hold. In Part II we apply this mathematical framework  to successfully model question order effects, response replicability and unpacking effects, thus opening the way toward quantum cognition beyond Hilbert space \cite{AertsSassolideBianchiSozzo2016}. 
\end{abstract}

\section{Introduction\label{intro}}
`Quantum cognition' is the name given to the approaches that successfully apply the mathematical formalism of quantum theory in Hilbert space to model cognitive phenomena. Conjunctive and disjunctive fallacies, conceptual over- and under-extensions, unpacking effects, and expected utility paradoxes are some of the  situations where quantum probabilistic approaches show significant advantages over the approaches in cognitive psychology that use classical Kolmogorovian probability theory (see, e.g., \cite{a2009a,bb2012,abgs2012,ags2012,pb2013,haven2013,WangEtal2014,AertsSozzoVeloz2015a,AertsSozzoVeloz2015b,busemeyernew}). Notwithstanding this success, two well known experimental situations, `question order effects' and `response replicability', seriously challenge the acceptance of Hilbert space  quantum cognition as a universally valid paradigm in cognitive psychology \cite{KhrennikovEtAl2014,KhrennikovBasieva2015,Boyer-KassemEtAl2015,AertsSassolideBianchi2015PlosOne}. We provide in Sec. \ref{challenges} an overview of the `no-go theorems' that Hilbert space imposes to the quantum representation of certain classes of psychological measurements.
 
The difficulties of quantum approaches to model the statistics of responses of sequential questions where these cognitive effects occur led us to investigate origins and range of applicability of the Born rule of quantum probability. To this end
we firstly developed an operational-realistic framework to describe cognitive entities, states and context-induced changes of states, individual and sequential measurements, measurement outcomes and their probabilities, etc. \cite{AertsSassolideBianchiSozzo2015}. This framework is applied in Sec. \ref{brussels} to the operational-realistic description of a wide class of dichotomic measurements, which includes those exhibiting the above mentioned cognitive effects. Next, we elaborated a `general tension-reduction (GTR) model' that extends some results obtained by some of us on the `hidden measurements interpretation of quantum probability' (see, e.g., \cite{Aertsetal1997}). The GTR-model, together with the associated `extended Bloch representation (EBR)', 
puts forward an explanation for the concrete effectiveness of the mathematical formalism of quantum theory in cognition. Indeed, the Born rule of quantum probability is characterized in it as a uniform fluctuation of the measurement context, and it emerges as a universal average over all possible forms of non-uniform fluctuations of the said context \cite{AertsSassolideBianchi2014,AertsSassolideBianchi2015a,AertsSassolideBianchi2015b}. In this way, the GTR-model is also able to explain the difficulties of the Hilbert space formalism in the simultaneous modeling of question order effects and response replicability within a quantum-like framework where the Born rule does not hold \cite{AertsSassolideBianchi2015PlosOne}.

We would like to point out at this stage a fundamental difference between quantum physics and quantum cognition, which we think is at the basis of the difficulties above. Consider a set of apparatuses performing measurements on a set of identically prepared physical entities. These apparatuses are different measurement contexts for the entities on which they act but, as long as they perform measurements of the same physical quantity, or observable, they are statistically indistinguishable in quantum physics. This entails that quantum probabilities can be indifferently obtained either by activating repeatedly the same measurement context, or by activating once a large number of different measurement contexts of the same quantum observable. The situation is different in a cognitive measurement, where  for a specific participant, acting as a context that performs a measurement on the cognitive entity under study, we can not know what his or her statistical identity is, when a first measurement takes place, because repeating the measurement will without doubt invoke context effect due to memories of the first measurements. Hence, in principle, different participants might carry different statistical identities even in their first measurement where memory does not yet come into play. However, even if this is the case, given the analysis in \cite{AertsSassolideBianchi2015a,AertsSassolideBianchi2015b} one can expect that if the sample of participants is large and diverse enough, the Born rule of quantum probability will arise naturally by averaging over all possible fluctuations of the measurement contexts determined by the participants. However, in the hypothesis that each individual carries a different statistical identity, if the sample of participants is not large enough or different measurements are performed sequentially on each participant, it might well be that the different measurement fluctuations cannot be averaged out in such a way that the Born rule is overall satisfied.

The latter is exactly the situation one has in question order effects and response replicability. We apply the GTR-model in Sec. \ref{GTRmodel} to represent dichotomic measurements that are performed individually 
and sequentially, together with the corresponding probabilities. We show that the GTR-model exhibits quantum-like aspects, but the Born rule of quantum probability is not generally valid in it, hence the GTR-model is generally non-Hilbertian. However, we also observe that the model is compatible with the operational and realistic framework for cognitive entities 
and we provide  an intuitive illustration of how it can be interpreted in cognition.

Hence, we present in Part I of this paper the fundamentals of a quantum-like approach for sequential measurements in cognition. The model in Sec. \ref{GTRmodel} is effectively applied in Part II of this paper to the experimental data collected in two experiments exhibiting question order effects. We show in the same paper that also response replicability requires a more general probabilistic framework in which the Born rule of quantum probability does not hold \cite{AertsSassolideBianchiSozzo2016}, and we explain how the unpacking effects can also be modeled and explained by the introduction of the non-Hilbertian but quantum-like probabilities of the GTR-model. The results of both Part I and Part II suggest that a more general quantum-like paradigm is needed in human cognition, which though goes beyond Hilbert space.

\section{Challenges to quantum cognition in Hilbert space\label{challenges}}
Notwithstanding its promising growth, quantum cognition in Hilbert space was recently challenged by two effects that can appear in certain classes of psychological measurements, namely, `question order effects'  and `response replicability' (the latter, however, is still waiting for a clear experimental confirmation). We illustrate them by using the definitions in \cite{KhrennikovEtAl2014}, as follows.

`Question order effects'. In an opinion poll, the response probabilities of two sequential questions depend on the order in which the questions are asked.

`Response replicability'. In an opinion poll, the response to a given question should give the same outcome if repeated, regardless of whether another question is asked and answered in between.

Let us consider an opinion poll and two dichotomic questions that are asked sequentially, in whatever order, on a sample of participants, such that probabilities of `yes' and `no' responses are collected as large number limits of statistical frequencies. The questions thus correspond to two `yes-no measurements' $A$ and $B$. The possible outcomes for $A$ and $B$ are `yes' and `no', which we denote by $A_{y}$ and $A_{n}$, and $B_{y}$ and $B_{n}$, respectively. Hence, performing first $A$ and then $B$ produces the possible outcomes $A_iB_j$, while performing first $B$ and then $A$ produces the possible outcomes $B_jA_i$, $i,j\in\{y,n\}$.

A question order effect occurs when, in a given cognitive situation, the probability distribution of measurement outcomes depends on the order in which the measurements are performed, i.e. $p(A_iB_j)\ne p(B_jA_i)$. Response replicability may instead appear in two forms,
`adjacent replicability' and `separated replicability' \cite{KhrennikovBasieva2015}. Suppose that the same measurement $A$ ($B$) is performed twice sequentially in a given cognitive situation. Then, adjacent replicability requires that, if the outcome $A_{i}$ ($B_{j}$) is obtained in the first measurement, then the same outcome $A_{i}$ ($B_{j}$) should be obtained in the second measurement with  certainty, i.e. probability 1. Suppose now that the sequence of measurements $ABA$ ($BAB$)
is performed in a given cognitive situation. Then, separated replicability requires that, if the outcome $A_{i}$ ($B_{j}$) is obtained in the first measurement, then the same outcome $A_{i}$ ($B_{j}$) should be obtained in the final measurement with certainty, i.e. probability 1. We thus formalize response replicability by $p(A_iA_i)=1$ in a 
$AA$
sequence, $p(B_jB_j)=1$ in a  
$BB$
sequence, $p(A_iB_jA_i)=1$ in a 
$ABA$
sequence, and $p(B_jA_iB_j)=1$ in a 
$BAB$
sequence. 

Let us now come to the way in which the above class of psychological measurements are modeled in Hilbert space. The cognitive situation is represented by a unit vector $|\psi\rangle$ of a Hilbert space, the measurements $A$ and $B$ are represented by the spectral measures $\{P_{i}^{A} \}$ and $\{P_{j}^{B} \}$, $i,j\in\{y,n\}$, and the Born rule is assumed to hold in both individual and sequential measurements, that is, $p_{\psi}(A_i)=\langle \psi|P_{i}^{A}|\psi\rangle$, $p_{\psi}(B_j)=\langle \psi|P_{j}^{B}|\psi\rangle$, $p_{\psi}(A_iB_j)=\langle \psi |P_{i}^{A}P_{j}^{B}P_{i}^{A}|\psi\rangle$, and $p_{\psi}(B_jA_i)=\langle \psi |P_{j}^{B}P_{i}^{A}P_{j}^{B}|\psi\rangle$, $i,j\in\{y,n\}$. Finally, this class of psychological measurements are assumed to be ideal first kind measurements in a standard quantum sense, hence the state transformations induced by the measurements $A$ and $B$  are $|\psi\rangle\to P_{i}^{A}|\psi\rangle/\| P_{i}^{A}|\psi\rangle \|$ and $|\psi\rangle\to P_{j}^{B}|\psi\rangle/\| P_{j}^{B}|\psi\rangle \|$, respectively, $i,j\in\{y,n\}$, according to L\"{u}ders postulate.

Deep studies confirm that, while the standard quantum formalism in Hilbert space is able to separately model question order effects and response replicability \cite{bb2012,pb2013,WangEtal2014,busemeyernew}, the same formalism does not work in cognitive situations where both effects are simultaneously present \cite{KhrennikovEtAl2014,KhrennikovBasieva2015,AertsSassolideBianchi2015PlosOne}. Roughly speaking, while the latter effect requires the spectral measures representing measurements to commute, the former can only be reproduced by non-commuting spectral families. The possibility of solving this problem by using more general positive operator values measurements (POVM) is still under investigation.

One may then wonder whether cognitive experiments exist where question order effects and response replicability are effectively observed. In this respect, one typically accepts the latter effect as a natural requirement for a wide class of psychological measurements. On the other hand, order effects in sequential measurements have been thoroughly studied since the seventies (see, e.g., \cite{SudmanBradburn1974,Moore2002}). In particular, Moore reviewed a Gallup poll conducted in 1997, in which he reported the results of different experiments on question order effects \cite{Moore2002}. Two of these experiments produced interesting results, namely, the `Clinton/Gore experiment'  and the `Rose/Jackson experiment'. 
Indeed, Hilbert space models of question order effects predict that a `QQ equality'
\begin{equation}
p_\psi(A_yB_y)-p_\psi(B_yA_y)+p_\psi(A_nB_n) -p_\psi(B_nA_n)=0
\end{equation}
should be satisfied by the experimental data, for every initial state $|\psi\rangle$ \cite{WangEtal2014}.

The QQ equality is important, as it provides a `parameter-free test of quantum models for question order effects'. Interestingly enough, this equality is approximately satisfied by the data collected in the Clinton/Gore experiment, while it is significantly violated in the Rose/Jackson experiment. Furthermore, some authors, including ourselves, lately observed that a special version of the quantum model, the `non-degenerate model', should  satisfy further parameter-free conditions, which are instead violated by the data \cite{Boyer-KassemEtAl2015,AertsSassolideBianchi2015PlosOne}.

We analyse in detail both the Clinton/Gore and Rose/Jackson experiments in \cite{AertsSassolideBianchiSozzo2016}  within our quantum-like GTR-model of sequential measurements. But, we can already draw a major result from the preceding discussion: at the level of question order effects, not only when the former are simultaneously present with response replicability, quantum modeling in Hilbert space is problematical, and a more general probabilistic framework becomes necessary.

\section{An operational-realistic framework for cognitive entities and measurements\label{brussels}}
In this section we apply the operational and realistic framework we developed in \cite{AertsSassolideBianchiSozzo2015} to the description of cognitive situations of the type presented in Sec.~\ref{challenges}. We will see in Part II of this paper that the perspective in this section is compatible with a quantum-like non-Hilbertian modeling of various cognitive effects, including question order effects, response replicability and unpacking effects \cite{AertsSassolideBianchiSozzo2016}.

Our framework rests on the operational and realistic foundations of quantum physics and quantum probability that were formalized by the SCoP formalism \cite{Aerts1999}. Here, the terms `operational' and `realistic' have a precise meaning. Our approach to cognition is `operational', i.e. the basic notions (states, measurements, outcomes and their probabilities, etc.) are defined in terms of the concrete operations that are performed in the laboratory of experimentation. Furthermore, our approach to cognition is `realistic', in the sense that the state of the cognitive entity is interpreted as a `state of affairs', hence it expresses a reality of the cognitive entity, albeit a reality not of a physical but of a conceptual nature. 

In experimental psychology, we can introduce `psychological laboratories', that is, spatio-temporal domains where cognitive experiments are performed. Let us focus ourselves on opinion polls, where a large number of human participants are asked questions in the form of structured questionnaires, and let the questions involve a `cognitive entity' $S$ (a concept, a combination of concepts, or a more complex conceptual situation). 

The experimental design, the questionnaire and the cognitive effect under study define a `preparation' of the cognitive entity $S$, which is thus assumed to be in an `initial state' ${p}_{S}$, and all participants interact with the cognitive entity in the same state $p_{S}$.
Suppose that the question, or yes-no measurement, $A$ is asked to a participant as part of the opinion poll. The measurement has the possible outcomes $A_{y}$ and $A_{n}$, depending on whether the response of the participant was `yes' or `no'. The interaction of the participant with the cognitive entity $S$ when the dichotomic measurement $A$ is performed leads to one of the two possible outcomes, and generally also  gives rise to a change of the state of the entity from $p_{S}$ to either $p_{A_y}$ or $p_{A_n}$, depending on whether the response is `yes' or `no'. Hence, the participant acts as a measurement context for the cognitive entity in the state $p_{S}$. If the same measurement $A$ is performed by making use of a large sample of participants, a statistics of responses is collected, which determines in the large number limit a `transition probability' $\mu(p_{A_{i}},e_{A},p_{S})$ that the initial state $p_{S}$ of the cognitive entity $S$ changes to the state $p_{A_{i}}$, $i\in\{y,n\}$, under the effect of the context $e_{A}$ determined by the measurement $A$.

The framework above formalizes the situation in Sec.~\ref{challenges}, where the participant is asked to answer `yes' or `no' to the question: ``Is Gore honest and trustworthy?''. If, for a given participant, the response is `yes', the  initial state  $p_{Honesty}$ of the conceptual entity {\it Honesty and Trustworthiness} (which we will denote {\it Honesty}, for the sake of simplicity) changes to a new state $p_{A_{y}}$, which is the state the entity is in when the choice `Gore is honest' is added to its original content.

Let us now suppose that a second question $B$ is asked to the participants as part of the opinion poll. This defines a measurement $B$, with possible outcomes $B_y$ and $B_n$, on the cognitive entity in the state $p_{S}$. Also in this case, the response determines a change of the state of $S$ from $p_{S}$ to either $p_{B_y}$ or $p_{B_n}$, depending on whether the response is `yes' or `no'. In the large number limit, we get a transition probability $\mu(p_{B_{j}},e_{B},p_{S})$ that the initial state $p_{S}$ of the cognitive entity $S$ changes to the state $p_{B_{j}}$, $j\in\{y,n\}$, under the effect of the context $e_{B}$ determined by the measurement $B$.

The framework above formalizes the situation in Sec.~\ref{challenges} where the participant is asked to answer `yes' or `no' to the question: ``Is Clinton honest and trustworthy?''. If, for a given participant, the response is `yes', the  initial state  $p_{Honesty}$ of the conceptual entity {\it Honesty and Trustworthiness} changes to a new state $p_{B_{y}}$, which is the state the entity is in when the choice `Clinton is honest' is added to its original content.

Then, let us suppose that each participant is first asked question $A$ and then question $B$. This defines a new measurement $AB$, with possible outcomes $A_iB_j$, $i,j\in\{y,n\}$, on the cognitive entity $S$ in the state $p_{S}$. The probability $p_{S}(A_iB_j)$ of obtaining the outcome $A_iB_j$ in the measurement $AB$, i.e. the outcome $A_i$ when performing $A$ and then $B_j$ when performing $B$, $i,j\in\{y,n\}$, on $S$ in the state $p_{S}$, is given by the product $p_{S}(A_iB_j)=\mu(p_{A_i},e_A,p_S)\mu(p_{B_j},e_B,p_{A_i})$.

Finally, let us suppose that each participant is first asked question $B$ and then question $A$. This defines a new measurement $BA$, with possible outcomes $B_jA_i$, $i,j\in\{y,n\}$, on the cognitive entity $S$ in the state $p_{S}$. The probability $p_{S}(B_jA_i)$ of obtaining the outcome $B_jA_i$ in the measurement $BA$, i.e. the outcome $B_j$ when performing $B$ and then $A_i$ when performing $A$, $i,j\in\{y,n\}$, on $S$ in the state $p_{S}$, is given by  the product $p_{S}(B_jA_i)=\mu(p_{B_j},e_B,p_S)\mu(p_{A_i},e_A,p_{B_j})$.

We stress, to conclude this section, that the state of a cognitive entity describes an element of a conceptual reality that is independent of any subjective belief of the person, or collection of persons, questioning about that entity. Such subjective beliefs are rather incorporated in the measurement context, which describes the cognitive interaction between the cognitive entity and the persons deciding on it. As such, our operational-realistic approach to cognition departs from other approaches that apply the quantum formalism to model cognitive phenomena \cite{bb2012,pb2013,haven2013}.  
 
\section{The GTR-model for dichotomic measurements\label{GTRmodel}}
We present here a geometric representation in the Euclidean 3-dimensional real space $\mathbb{R}^3$ of the operational and realistic entities we have introduced in Sec.~\ref{brussels}, focusing on the representation of the sequential measurements $AB$ and $BA$. Our results rest on \cite{AertsSassolideBianchi2015PlosOne}, to which we refer for technical details and calculations. The model presented here is an application of the GTR-model elaborated by ourselves, where quantum probabilities are recovered as universal averages over all possible forms of non-uniform fluctuations \cite{AertsSassolideBianchi2015a,AertsSassolideBianchi2015b}. When the structure of the state space is  Hilbertian, as in quantum physics, the GTR-model  
reduces to the so-called  `extended Bloch representation (EBR)'  of quantum theory \cite{AertsSassolideBianchi2014}.

Let us firstly consider individual measurements with two outcomes on a cognitive entity and study how they are represented in the  EBR representation.

\begin{figure}[!ht]
\centering
\includegraphics[scale =.28]{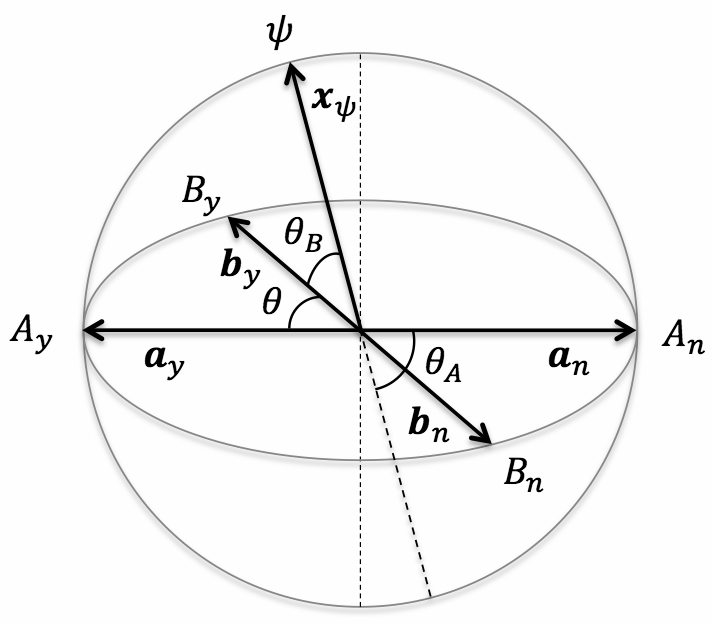}
\caption{A representation of the initial state of the cognitive entity $S$, and of the outcome states of the $A$ and $B$ measurements, on the Bloch sphere.\label{Bloch}}
\end{figure}

The cognitive entity $S$ is represented by an abstract point particle that can move on the surface of a 3-dimensional unit sphere, called 
the `Bloch sphere'. The initial state $p_{S}$ of $S$ is represented by a state of the point particle on the sphere corresponding to the position ${\bf x}_\psi$ on the sphere (see Fig.~ \ref{Bloch}). 
Dichotomic measurements on $S$ are represented by $1$-dimensional breakable and elastic structures, anchored at two antipodal points, corresponding to the two possible  outcome states. More precisely, the measurement $A$ is represented by a breakable elastic band stretched between the two points ${\bf a}_y$ and ${\bf a}_n = -{\bf a}_y$, $\|{\bf a}_y\| = \|{\bf a}_n\|=1$, corresponding to the two outcomes $A_{y}$ and $A_{n}$, respectively. Analogously, the measurement $B$ is represented by a breakable elastic band stretched between the two points ${\bf b}_y$ and ${\bf b}_n = -{\bf b}_y$, $\|{\bf b}_y\| = \|{\bf b}_n\|=1$, corresponding to the two outcomes $B_{y}$ and $B_{n}$, respectively. Equivalently, the states $p_{A_i}$ and $p_{B_j}$ in Sec. \ref{brussels} are represented by the positions ${\bf a}_i$ and ${\bf b}_j$, respectively, $i,j\in\{y,n\}$.

We assume that the two breakable elastics are parameterized in such a way that the coordinate $x=1$ ($x=-1$) corresponds to the outcome `yes' (`no'), where $x=0$ describes the center, which also coincides with the center of the Bloch sphere. Each elastic represents a possible dichotomic measurement, and is described not only by its orientation within the sphere, but also by `the way' it can break. More concretely, breakability of the elastic representing the measurement $A$ is formalized by a probability distribution $\rho_A(x|\psi)$ such that $\int_{x_1}^{x_2}\rho_A(x|\psi)dx$ is the probability that the elastic breaks in a point in the interval $[x_1,x_2]$, $-1\leq x_1\leq x_2\leq 1$, when the measurement $A$ is performed and the point particle is in the initial position ${\bf x}_\psi$. The condition $\int_{-1}^{1}\rho_A(x|\psi)dx=1$ guarantees that the elastic will break in one of its points, with 
certainty, i.e. that the measurement will produce an outcome.

\begin{figure}[!ht]
\centering
\includegraphics[scale =.28]{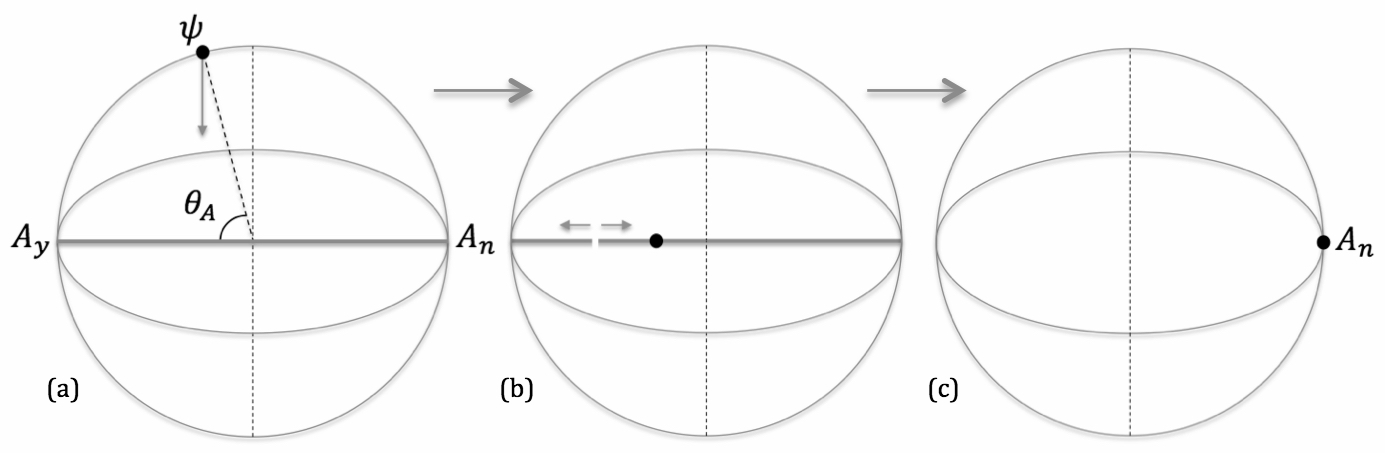}
\caption{A description of the unfolding of the $A$-measurement process within the Bloch sphere: (a) the point particle orthogonally ``fall'' onto the elastic band; (b) the elastic then breaks at some unpredictable point; (c) the point particle is finally drawn to one of the two elastic's end points, here $A_n$.
\label{Measurement}}
\end{figure}

Let us now describe the measurement $A$ on the cognitive entity $S$ in the state $p_S$ as represented  in the Bloch sphere. When the measurement $A$ is performed and the point particle is in the initial  position ${\bf x}_\psi$ on the Bloch sphere, a certain probability distribution $\rho_A(x|\psi)$ is
actualized, which describes the way the $A$-elastic band will break, in accordance with the fluctuations that are present in the measurement context $e_{A}$. Then, the point particle  ``falls'' from its original position ${\bf x}_\psi$ orthogonally onto the $A$-elastic band and sticks to it. Next, the elastic breaks in some point, and its two broken fragments contract toward the corresponding anchor points, bringing with them the point particle (Fig. \ref{Measurement}). 

If $x_A$ is the position of the point particle onto the elastic, i.e. $x_A= {\bf x}_\psi \cdot {\bf a}_y =\cos\theta_A$, and the elastic breaks in a point $\lambda$, with $x_A<\lambda$, then the particle attached to the elastic fragment $[-1,\lambda]$ is drawn toward the position ${\bf a}_y$. In this case, we say that the measurement $A$ gives  the outcome `yes'. If instead $x_A>\lambda$, then the particle attached to the elastic fragment  $[\lambda, 1]$
is drawn toward the position ${\bf a}_n$. In this case, we say that the measurement $A$ gives  the outcome `no'.\footnote{If $\lambda = \cos\theta_A$, we are in a situation of classical unstable equilibrium, and the outcome is not predetermined. However, these  exceptional values of $\lambda$ have zero measure, hence they do not contribute to the probabilities in (\ref{A-integrals}).} 
The transition probabilities, $p_{\psi}(A_y)$ that the  initial  position ${\bf x}_\psi$ collapses to ${\bf a}_y$, and $p_{\psi}(A_n)$ that the  initial 
position ${\bf x}_\psi$ collapses to ${\bf a}_n$, are given by
\begin{equation}
p_{\psi}(A_y)=\int_{-1}^{\cos\theta_A}\rho_A(x|\psi) dx  \quad\quad p_{\psi}(A_n)=\int_{\cos\theta_A}^{1}\rho_A(x|\psi) dx
\label{A-integrals}
\end{equation}
and represent the transition probabilities $\mu(p_{A_y},e_A,p_{S})$ and $\mu(p_{A_n},e_A,p_{S})$, respectively, 
introduced in Sec.~\ref{brussels}.

It is worth noticing that: (i) the probabilities in (\ref{A-integrals}) formalize a lack of knowledge about the measurement process, i.e. the breaking point $\lambda$ corresponds to a  `hidden measurement-interaction'; (ii) the Born rule of quantum probability is recovered when $\rho_A(x|\psi)=\frac{1}{2}$, i.e. the probability distribution is globally uniform, in which case (\ref{A-integrals}) becomes
\begin{equation}
p_{\psi}(A_y)={1\over 2}(1+\cos\theta_A) \quad\quad p_{\psi}(A_n)={1\over 2}(1-\cos\theta_A).
\label{A-quantum}
\end{equation}
This result is not limited  to dichotomic measurements, but has a general validity, i.e. it can be naturally generalized to degenerate and non-degenerate measurements having an arbitrary number of outcomes \cite{AertsSassolideBianchi2014,AertsSassolideBianchi2015a,AertsSassolideBianchi2015b}.

Let us now come to the  measurement $B$. Proceeding as above, we have that the transition probabilities, $p_{\psi}(B_y)$ that the position ${\bf x}_\psi$ collapses to ${\bf b}_y$ (`yes' outcome is obtained), and $p_{\psi}(B_n)$ that the position ${\bf x}_\psi$ collapses to ${\bf b}_n$
(the outcome `no' is obtained),  are given by
\begin{equation}
p_{\psi}(B_y)=\int_{-1}^{\cos\theta_B}\rho_B(x|\psi) dx \quad\quad p_{\psi}(B_n)=\int_{\cos\theta_B}^{1}\rho_B(x|\psi) dx
\label{B-integrals}
\end{equation}
and  represent the transition probabilities $\mu(p_{B_y},e_B,p_{S})$ and $\mu(p_{B_n},e_B,p_{S})$, respectively, introduced  in Sec. \ref{brussels}. 
In (\ref{B-integrals}), $x_B= {\bf x}_\psi\cdot {\bf b}_y=\cos\theta_B$ denotes the landing point of the point particle onto the $B$-elastic band, while $\rho_B(x|\psi)$ denotes the probability distribution associated with the latter. 

Let us then consider sequential measurements on a cognitive entity and study how they are represented in the GTR-model. Suppose that we  
firstly perform the measurement $A$ and then the measurement $B$. We then have the four transition probabilities $p_{\psi}(A_iB_j)$ that the 
point particle position ${\bf x}_\psi$,  representing the initial state, first changes to the position ${\bf a}_i$ and then to the position ${\bf b}_j$ (sequential outcome $A_i$ and then $B_j$), $i,j\in\{y,n\}$. If we set $\cos\theta = {\bf a}_y\cdot {\bf b}_y$ (see Fig.~\ref{Bloch}), 
we can first write the conditional probabilities $p_{A_i}(B_j)$ that the position ${\bf a}_i$ changes to the position ${\bf b}_j$, $i,j\in\{y,n\}$, as 
\begin{eqnarray}
&&p_{A_y}(B_y)=\int_{-1}^{\cos\theta}\rho_B(x| A_y) dx \quad\quad p_{A_y}(B_n)=\int_{\cos\theta}^{1}\rho_B(x| A_y) dx \nonumber\\
&&p_{A_n}(B_y)=\int_{-1}^{-\cos\theta}\rho_B(x| A_n) dx \quad\quad p_{A_n}(B_n)=\int_{-\cos\theta}^{1}\rho_B(x| A_n) dx
\label{B|A-integrals}
\end{eqnarray}
where $\rho_B(x|A_y)$ (respectively $\rho_B(x|A_n)$) is the probability distribution actualized during the measurement $B$, knowing that the measurement $A$ produced the transition from  ${\bf x}_\psi$ to ${\bf a}_y$ (respectively to ${\bf a}_n$).

Now, for every $i,j\in\{y,n\}$, we have  $p_{\psi}(A_iB_j)=p_{\psi}(A_i)p_{A_i}(B_j)$ 
for the transition probabilities in the sequential measurement $AB$. More explicitly, using (\ref{B|A-integrals}) and (\ref{B-integrals}), we can write:
\begin{eqnarray}
&&p_{\psi}(A_yB_y)= \int_{-1}^{\cos\theta}\rho_B(x| A_y) dx \int_{-1}^{\cos\theta_A}\rho_A(x|\psi) dx\nonumber\\
&&p_{\psi}(A_yB_n)= \int_{\cos\theta}^{1}\rho_B(x| A_y) dx \int_{-1}^{\cos\theta_A}\rho_A(x|\psi) dx\nonumber\\
&&p_{\psi}(A_nB_y)= \int_{-1}^{-\cos\theta}\rho_B(x| A_n) dx \int_{\cos\theta_A}^{1}\rho_A(x|\psi) dx\nonumber\\
&&p_{\psi}(A_nB_n)=\int_{-\cos\theta}^{1}\rho_B(x| A_n) dx \int_{\cos\theta_A}^{1}\rho_A(x|\psi) dx.
\label{AB-integrals}
\end{eqnarray}
By exchanging the role of  $A$ and $B$ in (\ref{AB-integrals}), we get the following similar expressions for the probabilities of the sequential measurement $BA$:
\begin{eqnarray}
&&p_{\psi}(B_yA_y)= \int_{-1}^{\cos\theta}\rho_A(x|B_y) dx \int_{-1}^{\cos\theta_B}\rho_B(x|\psi) dx,\nonumber\\
&&p_{\psi}(B_yA_n)= \int_{\cos\theta}^{1}\rho_A(x|B_y) dx \int_{-1}^{\cos\theta_B}\rho_B(x|\psi) dx,\nonumber\\
&&p_{\psi}(B_nA_y)= \int_{-1}^{-\cos\theta}\rho_A(x|B_n) dx \int_{\cos\theta_B}^{1}\rho_B(x|\psi) dx,\nonumber\\
&&p_{\psi}(B_nA_n)=\int_{-\cos\theta}^{1}\rho_A(x|B_n) dx \int_{\cos\theta_B}^{1}\rho_B(x|\psi) dx.
\label{BA-integrals}
\end{eqnarray}
Clearly,  the probabilities (\ref{AB-integrals}) and (\ref{BA-integrals}) coincide by construction with the sequential probabilities $p_{S}(A_iB_j)$ and $p_{S}(B_jA_i)$, respectively, $i,j\in\{y,n\}$, given in Sec.~\ref{brussels}.

Our general modeling of cognitive entities, states, dichotomic measurements and sequential measurement processes is thus completed. One realizes at once that it incorporates quantum aspects, as context induced changes of state, pure potentiality, unavoidable and uncontrollable uncertainty. In this sense, one can say that the model of dichotomic sequential measurements that we have presented  is quantum-like. However, it is much more general than the standard Hilbert space representation, as the Born rule of quantum probability is only recovered in the specific case in which $\rho_A$ and $\rho_B$ are both globally uniform probability distributions (describing uniform elastic structures, having the same probability to break in all their points). 

In order to find explicit solutions, to be used in specific applications,  one needs to add some reasonable constraints to the measurements $A$ and $B$, in particular for what concerns the probability densities $\rho_A$ and $\rho_B$. This task is accomplished in Part II of this paper, where we apply the quantum-like modeling presented here to faithfully  represent the experimental data on the question order effects identified in the Clinton/Gore and Rose/Jackson experiments, as well as to represent response replicability and unpacking effects  \cite{AertsSassolideBianchiSozzo2016}. We instead conclude Part I of this paper by providing an intuitive illustration of how an elastic model for sequential measurements can be interpreted in cognition.

The elastic mechanism described in Sec.~\ref{GTRmodel} also provides a possible representation of what we intuitively feel when confronted with decisional contexts, and a neural/mental equilibrium is progressively built, resulting from the balancing of the different tensions between the initial state and the available mutually excluding answers. Indeed, an elastics stretched between two antipodal points in the Bloch sphere can be seen as an abstract representation of such equilibrium, which at some moment will be altered in a non-predictable way (when the elastic breaks), causing a sudden and irreversible process during which the initial conceptual state is drawn to one of the possible answers.

The compatibility of the GTR-model with our intuitive understanding of the human cognitive processes remains such also when psychological measurements with an arbitrary number $N$ of outcomes are considered \cite{AertsSassolideBianchi2014,AertsSassolideBianchi2015a,AertsSassolideBianchi2015b}. The elastics are then replaced by disintegrable hyper-membranes having the shape of $(N-1)$-dimensional simplexes. Similarly to the $N=2$ situation, the latter can still be viewed not only as mathematical objects naturally representing the measurements' probabilities, and their relations, but also as a way to `give shape' to the different mental states of equilibrium, characterized by the existence of different competing `tension lines' going from the on-membrane position of the point particle to the $N$ vertices of the simplex, representing the different answers. These `tension-reduction processes' can also describe situations where the conflicts between the competing answers cannot be fully resolved, so that the system is brought into another state of equilibrium, between a reduced set of possibilities, which in the GTR-model correspond to lower-dimensional sub-simplexes \cite{AertsSassolideBianchi2014,AertsSassolideBianchi2015a,AertsSassolideBianchi2015b}. These are situations describing sub-measurements of a given mesurement, called degenerate measurements in quantum mechanics. As we shall see in Part II \cite{AertsSassolideBianchiSozzo2016}, they may have some relevance in the description of the unpacking effects.

\end{document}